\documentclass[11pt]{article}

\usepackage[preprint]{acl}

\usepackage{times}
\usepackage{latexsym}

\usepackage[T1]{fontenc}

\usepackage[utf8]{inputenc}

\usepackage{microtype}

\usepackage{inconsolata}

\usepackage{graphicx}

\usepackage{booktabs}
\usepackage{multirow}
\usepackage{amsmath}
\usepackage{comment}
\usepackage{amsfonts, amssymb}
\usepackage{siunitx}
\usepackage{makecell}

\title{From Self to Other: Evaluating Demographic Perspective-Taking in LLM Hate Speech Annotation}

\author{Paloma Piot \\
  Information Retrieval Lab \\ CITIC Research Centre \\ Universidade da Coruña \\
  \texttt{paloma.piot@udc.es} \\\And
  Javier Parapar \\
  Information Retrieval Lab \\ CITIC Research Centre \\ Universidade da Coruña  \\
  \texttt{javier.parapar@udc.es} \\}

\begin{document}
\maketitle
\begin{abstract}
Hate speech detection is inherently subjective: people from different demographic groups perceive the same content very differently. Collecting enough annotations from multiple demographic groups is costly and difficult to scale. Persona-conditioned Large Language Models (models prompted to adopt a specific demographic identity) have been proposed as a way to simulate diverse perspectives at scale. But do they actually reflect how different groups disagree? We evaluate three aspects of human social judgement: $(i)$ whether personas from different groups disagree in human-like ways (inter-group disagreement), $(ii)$ whether they become more sensitive when content targets their own identity (in-group sensitivity), and $(iii)$ whether they can accurately predict how another group would react (vicarious prediction). Our results show that no model consistently captures all three dimensions, and performance is highly model-dependent and does not emerge reliably from minimal identity prompts alone. However, vicarious prompting with \texttt{Llama~3.1} yields the highest cross-group agreement in most demographic axes and provides the closest overall approximation to human disagreement patterns, indicating that this configuration may provide a more reliable setting for automatic annotation aligned with human judgements.\\
\textcolor{red}{This article contains illustrative instances of hateful language.}
\end{abstract}

\section{Introduction}
\label{sec:introduction}
The perception of hate speech is inherently subjective: what counts as offensive or harmful depends on the observer's social identity, lived experience, and normative expectations \cite{rottger-etal-2022-two, sap-etal-2022-annotators, davani-etal-2023-hate, dehghan2025dealingannotatordisagreementhate}. Disagreement between annotators from different demographic groups is not mere noise; it reflects genuine diversity in interpretation \cite{10.1007/978-3-031-08974-9_54, davani-etal-2022-dealing, barbarestani-etal-2024-content}. In practice, content moderation often collapses these diverse viewpoints into a single ground-truth label, risking systematic under-enforcement or over-enforcement for certain communities and raising concerns about fairness, accountability, and the inclusivity of automated systems \cite{davani-etal-2023-hate, Cabitza_Campagner_Basile_2023, moscato-etal-2025-personalization}.


At the same time, Large Language Models (LLMs) are increasingly used as proxy annotators or social simulators \cite{Gilardi2023}. Persona conditioning, that is, prompting models to adopt specific demographic identities, has been proposed to recover subjectivity by simulating different perspectives \cite{10.5555/3618408.3619652, Argyle_Busby_Fulda_Gubler_Rytting_Wingate_2023, plaza-del-arco-etal-2024-divine, Piot_Martin-Rodilla_Parapar_2026}. While prior work evaluates these models by comparing label alignment with human averages, such aggregate metrics overlook the social structure of judgement: how groups differ in targeted content and how they perceive other groups' reactions.

In this work, we assess whether zero-shot persona-conditioned LLMs can provide practically useful insights for annotation and moderation by capturing three dimensions of human social judgement: $(i)$ First, we examine whether LLM personas reproduce the patterns of disagreement observed between human demographic groups when judging hate speech. $(ii)$ Second, we test whether annotator behaviour changes when comments target their own identity group and whether LLM personas show similar shifts. $(iii)$ Third, we evaluate whether LLM personas change their judgements when prompted to adopt a vicarious perspective and predict how another demographic group would rate the same content. By evaluating these dimensions, we examine whether persona-conditioned LLMs capture key patterns of human social judgement and whether they could serve as a useful proxy for demographic perspectives in content annotation and moderation.

To do so, we address the following research questions:
\begin{itemize}
    \item \textbf{RQ1 (Inter-group disagreement)}: \textit{Do LLM personas replicate patterns of disagreement observed between human demographic groups when judging hate speech?}

    \item \textbf{RQ2 (In-group sensitivity)}: \textit{Do annotators judge hate speech differently when their own identity group is targeted, and do LLM personas exhibit similar patterns relative to human annotations?}
    
    \item \textbf{RQ3 (Out-group perception)}: \textit{Do LLM personas change their hate speech judgements under vicarious prompting (i.e., when evaluating from another group’s perspective), and how do these predictions align with human consensus?}
\end{itemize}

Our results show differences across the three research questions. For \textbf{RQ1} (inter-group disagreement), \texttt{Nemo} most closely approximates human cross-group agreement levels, while \texttt{Llama~3.1} diverges strongly and \texttt{Qwen~3} shows weak alignment with human disagreement patterns. For \textbf{RQ2} (in-group sensitivity), \texttt{Llama~3.1} partially reproduces the human pattern of judgement shifts when comments target an annotator's own identity group, whereas \texttt{Nemo} and \texttt{Qwen~3} show little or inconsistent alignment. For \textbf{RQ3} (out-group perception), \texttt{Llama~3.1} exhibits consistent effects under vicarious prompting, while \texttt{Nemo} shows almost no change in its judgements, and \texttt{Qwen~3} again shows weak effects. Taken together, these findings indicate that judgement patterns produced by persona-conditioned LLMs are highly model-dependent and do not reliably emerge from minimal identity prompts alone. While vicarious prompting improves alignment with human cross-group annotations in our experiments, persona outputs should not be assumed to faithfully represent the perceptual diversity of human annotators without careful empirical validation.

\section{Related Work}

\paragraph{Subjectivity and Disagreement in NLP.} 
Perspectivism in NLP argues that language understanding is inherently subjective and that multiple valid interpretations should be preserved rather than collapsed into a single ground truth~\cite{10.1007/978-3-030-77091-4_26}. \citet{davani-etal-2022-dealing} demonstrate that annotator disagreement encodes meaningful social and cultural variation rather than noise. This has led to a shift toward modelling annotator-specific labels, distributions, or latent perspectives, treating disagreement as a primary signal in NLP systems~\cite{barbarestani-etal-2024-content}. Recent work corroborates this view empirically, showing that high annotator disagreement meaningfully signals cases where subjective factors shape toxicity perception, rather than reflecting inconsistency~\cite{pachinger-etal-2025-disaggregated}. However, \citet{orlikowski-etal-2023-ecological} caution that annotation variation cannot be explained by group-level sociodemographics alone: demographic attributes do not significantly improve models of individual annotator behaviour, suggesting that identity cues capture only part of the perceptual structure underlying disagreement.

\begin{table*}[t]
    \centering
    \small
    \begin{tabular}{p{3.4cm} p{4.3cm} p{7cm}}
    \toprule
    \textbf{Experiment} & \textbf{Subset Requirement} & \textbf{Purpose} \\
    \midrule
    \textbf{RQ1}: Inter-group disagreement (Table~\ref{tab:data_rq1_overlap_comments})
    & Comments annotated by \textbf{both} demographic groups 
    & Enables direct comparison of group annotations to measure inter-group agreement \\
    \midrule
    \textbf{RQ2}: In-group sensitivity (Table~\ref{tab:data_rq2_targeted_comments})
    & Comments \textbf{targeting} a specific identity group 
    & Evaluates whether annotators respond differently when their own group is targeted vs. non-targeted comments \\
    \midrule
    \textbf{RQ3}: Vicarious prediction (Table~\ref{tab:data_rq3})
    & Predictions by the LLM persona in self vs. \textbf{vicarious out-group} mode
    & Measures alignment of vicarious predictions with both the out-group's and global human majority \\
    \bottomrule
    \end{tabular}
    \caption{Dataset configurations used for each research question. Each experiment requires a different subset of the data depending on the annotation structure and evaluation objective.}
    \label{tab:dataset_subsets}
\end{table*}

\paragraph{LLMs for Data Annotation \& Simulation.}
LLMs are increasingly used as substitutes for human annotators, often approximating crowd judgements across classification tasks~\cite{10.1007/978-3-031-78548-1_2, Gilardi2023}. However, models may also exhibit cognitive biases such as the False Consensus Effect, overestimating how closely their judgements align with those of others~\cite{choi-etal-2025-people}. Early work suggested that LLMs could function as ``silicon samples'' of human subpopulations, showing that conditioning GPT-3 on sociodemographic backstories can produce response distributions resembling those of particular groups~\cite{Argyle_Busby_Fulda_Gubler_Rytting_Wingate_2023}. However, subsequent research has questioned the reliability of this simulation paradigm. While LLMs can sometimes preserve relative performance trends when evaluating models, they often diverge from humans at the instance level, suggesting that they may function better as proxy evaluators than as direct annotator replacements~\cite{piot-etal-2026-can}. Similarly, large-scale benchmarks of human behaviour simulation show that even state-of-the-art models achieve limited fidelity, particularly on subjective or high-entropy questions and when simulating groups defined by religion or ideology~\cite{hu2025simbenchbenchmarkingabilitylarge}. Safety alignment mechanisms such as RLHF may suppress controversial responses and reduce output diversity~\cite{rottger-etal-2024-xstest}, while demographic alignment appears to depend more on task difficulty than on explicit identity prompting~\cite{brown2025evaluatingllmannotationsrepresent}. More broadly, annotation outputs vary substantially across model choices and prompting strategies, sometimes leading to incorrect scientific conclusions, a phenomenon termed \textit{LLM hacking}~\cite{baumann2025largelanguagemodelhacking}.

\paragraph{Algorithmic Bias \& Persona Stereotyping.}
Recent studies indicate that persona-conditioned LLMs can produce inconsistent or stereotypical outputs, which limits their validity as proxies for human demographic groups \cite{10.5555/3618408.3619652}. \citet{Piot_Martin-Rodilla_Parapar_2026} demonstrate that country specific persona prompts can significantly alter hate speech classification outcomes, showing that model judgements are sensitive to identity framing. Additionally, work on vicarious offence examines how both humans and models predict the reactions of others, finding systematic biases in how out-group sensitivity is estimated \cite{weerasooriya-etal-2023-vicarious}.

\section{Methodology}
\label{sec:methodology}


\subsection{Data}
\label{subsec:data}

We use the \textit{Measuring Hate Speech} (MHS) dataset \cite{kennedy2020constructing}, which contains over 135,000 crowd-sourced annotations for nearly 40,000 comments sourced from YouTube, Reddit, and Twitter. The dataset includes detailed demographic metadata on gender, race, religion, ideology, and sexuality. Each comment is rated on a three-point scale: 0 $=$ no hate, 1 $=$ unsure, and 2 $=$ hate. To focus on clear judgement differences rather than ambiguity in the annotation process, we remove annotations labelled as ``unsure'' (1) and retain only the binary labels 0 and 2. This allows reliable estimation of inter-group disagreement, in-group sensitivity, and vicarious perception, making MHS the most suitable resource for evaluating whether persona-conditioned LLMs capture relative differences in human perception. Table~\ref{tab:full_demographics} (Appendix) provides the demographic distribution of annotators and the number of comments rated by each group. While MHS is English-only and its multi-platform domain composition (YouTube, Reddit, Twitter) presents specific community norms that may constrain broad generalizability, it remains the best available dataset for our study, as most existing hate speech corpora lack sufficient demographic information for meaningful multi-axis comparisons.

\paragraph{Demographic selection.}
We analyse five primary binary demographic pairs, selected to maximise statistical power by focusing on the groups with the highest annotation density in the dataset. These axes also correspond to dimensions where disagreement in hate speech perception is commonly observed~\cite{sap-etal-2022-annotators}.

\begin{itemize}
    \item \textbf{Gender:} Men vs. Women (non-binary annotators excluded due to limited sample size).
    \item \textbf{Race:} White vs. Black.    
    \item \textbf{Religion:} Christian vs. Atheist.
    \item \textbf{Ideology:} Liberal vs. Conservative (comparing only the ideological extremes).
    \item \textbf{Sexuality:} Straight vs. Gay.
\end{itemize}

Because each research question probes a distinct dimension of model–human behaviour, we evaluate it on the subset of comments and annotators most relevant to that dimension (see Appendix Tables~\ref{tab:data_rq1_overlap_comments},~\ref{tab:data_rq2_targeted_comments},~\ref{tab:data_rq3}). Table~\ref{tab:dataset_subsets} summarises these subsets.

\subsection{Models}
\label{subsec:models}
To test whether persona-conditioned LLM prediction patterns might generalise across training distributions and alignment regimes, we evaluate three open-weight, state-of-the-art models. We focus on open models for reproducibility and transparency, selecting a single mid-size variant (8B–14B parameters) per family to balance computational efficiency with sufficient reasoning capacity. Closed-source models are outside the scope of this study due to the lack of transparency regarding their training data and alignment processes.

\begin{enumerate}
    \item \textbf{\texttt{Llama~3.1 8B}}~\cite{grattafiori2024llama3herdmodels}: Developed by Meta, this model is trained on a large multilingual corpus of publicly available web data, with English comprising the majority of the training tokens.
    \item \textbf{\texttt{Nemo 12B}}~\cite{mistral2024nemo}: Developed by Mistral AI in collaboration with NVIDIA, this model is trained on a multilingual corpus and designed to support a wide range of languages and programming tasks.
    \item \textbf{\texttt{Qwen~3 14B}}~\cite{bai2023qwentechnicalreport}: Developed by Alibaba Cloud, this model is trained on a large multilingual corpus spanning over 100 languages, reflecting a broad global data distribution.
\end{enumerate}

These models have been shown to exhibit distinct political orientations that vary across languages and scales \cite{gurgurov-etal-2025-multilingual}, further motivating our choice to evaluate culturally diverse model families.


\subsection{Prompting Strategy}
\label{subsec:prompting}
We intentionally adopt a zero-shot prompting strategy to isolate the models' intrinsic representations of social groups and hate speech perception. Unlike few-shot or instruction-tuned models, this approach avoids external guidance that could steer the model, allowing us to evaluate whether persona-conditioned behaviours emerge from pre-training and alignment alone. Thus, our findings reflect the baseline social reasoning capacity of the models rather than their upper-bound performance under prompt engineering.

To ensure consistency across demographic groups, we use a simple, uniform persona template (e.g., ``\textit{Adopt the identity of [Demographic A]}'') following prior work~\cite{gupta2024biasrunsdeepimplicit}. This minimal specification avoids unequal or hand-crafted descriptions that could bias behaviour while allowing us to test whether models activate coherent social representations from identity cues alone. Richer persona descriptions (e.g., including cultural or experiential details) may yield stronger simulations but risk introducing researcher-defined assumptions and are left for future work.

We evaluate two prompting modes: \textbf{self-alignment}, where the model rates a comment from the perspective of its demographic group (e.g., women evaluating from a women's perspective), and \textbf{vicarious}, where the model predicts the perception of an out-group demographic (e.g., women predicting how men would perceive the comment). The full prompt templates are provided in Appendix~\ref{app:prompts}.

\subsection{Evaluation Metrics}
\label{subsec:metrics}

We evaluate human annotators and LLM personas using three metrics corresponding to our research questions: \textbf{RQ1} inter-group disagreement, \textbf{RQ2} in-group sensitivity, and \textbf{RQ3} out-group prediction shift.

\subsubsection{RQ1: Inter-group Disagreement}
\label{subsubsec:disagreement_metrics}
To quantify disagreement between demographic groups $(A, B)$, we compute Cohen's $\kappa$ between their label sequences. For humans, each comment is assigned the majority vote per group (e.g., majority vote of women vs. majority vote of men); for LLMs, each persona provides a single label per comment (e.g., one prediction label for women vs. one prediction label for men). Low $\kappa$ indicates disagreement, while values near 1 indicate high agreement. This metric assesses how well persona-conditioned LLMs replicate patterns of disagreement observed in human annotators.

\subsubsection{RQ2: In-group Sensitivity}
\label{subsec:in-group-sensitivity}

In-group sensitivity measures whether a group's judgements differ when evaluating comments that target their own identity versus comments targeting others. We define two conditions: $(i)$~\textbf{Targeted}: comments explicitly targeting the focal group; $(ii)$~\textbf{Non-targeted}: all other comments.  

For both humans and LLMs, sensitivity is computed using Cohen's $\kappa$ against the majority vote of annotators outside the focal group ($\neg A$). For humans, this compares group $A$'s majority with $\neg A$; for LLMs, it compares the persona's prediction with $\neg A$’s majority.  

To quantify how the presence of a target group affects agreement, we compute $\Delta_{\text{IG}}$ as:

\[
    \Delta_{\text{IG}}(A) = 
    \kappa\bigl(\hat{y}_A^{\text{targ}},\ \hat{y}_{\neg A}^{\text{targ}}\bigr) -
    \kappa\bigl(\hat{y}_A^{\text{no-targ}},\ \hat{y}_{\neg A}^{\text{no-targ}}\bigr)
\]

A positive $\Delta_{\text{IG}}$ indicates that agreement with out-group annotators increases for comments targeting the group itself, reflecting greater in-group sensitivity; a value near zero indicates no change depending on the comment's target.


To assess statistical significance, we apply Welch's independent-samples $t$-test to the comment-level binary agreement scores underlying the reported $\kappa$ values, as the targeted and non-targeted comment subsets are mutually exclusive and therefore treated as independent. We confirm robustness using the non-parametric Mann--Whitney $U$ test, which does not assume normality and is appropriate for binary outcomes. All $p$-values are corrected for multiple comparisons using the Benjamini–Hochberg procedure.

\subsubsection{RQ3: Out-group Prediction}
\label{subsec:vicarious}
To evaluate whether a persona can predict another group's perspective, we compare the persona's self-predictions $\hat{y}^{\text{self}}_A$ with vicarious predictions $\hat{y}^{\text{vic}}_{A \rightarrow B}$ of an out-group $B$.  Agreement with the out-group majority $\hat{y}_B$ is measured using Cohen’s $\kappa$:

\[
\kappa^{\text{self}} = \kappa(\hat{y}^{\text{self}}_A, \hat{y}_B), \quad
\kappa^{\text{vic}} = \kappa(\hat{y}^{\text{vic}}_{A \rightarrow B}, \hat{y}_B)
\]

We additionally compute alignment with the global majority vote $\hat{y}$:

\[
\kappa^{\text{self, all}} = \kappa(\hat{y}^{\text{self}}_A, \hat{y}), \quad
\kappa^{\text{vic, all}} = \kappa(\hat{y}^{\text{vic}}_{A \rightarrow B}, \hat{y})
\]

The \textit{vicarious gap} quantifies the effect of prompting the persona to adopt the out-group perspective:

\[
\Delta_{A \rightarrow B} = \kappa^{\text{vic}} - \kappa^{\text{self}}
\]

A positive $\Delta_{A \rightarrow B}$ indicates that the vicarious prompt improves alignment with group $B$’s labels, while a value near zero indicates little or no change, and a negative value indicates reduced alignment. Similarly, positive values of $\kappa^{\text{vic, all}} - \kappa^{\text{self, all}}$ indicate improved alignment with the overall human consensus. To test whether $\Delta_{A \rightarrow B}$ differs significantly from zero, we apply a one-sample $t$-test to the comment-level difference between vicarious and self predictions, as the data consist of paired observations whose mean difference can be evaluated against zero. Since the difference scores are discrete and may deviate from normality, we also report the Wilcoxon signed-rank test as a non-parametric alternative. All $p$-values are corrected for multiple comparisons using the Benjamini--Hochberg procedure.

\section{Experiments}
\label{sec:experiments}
We report results for the five demographic comparison pairs across the three model families described in Sections~\ref{subsec:data} and~\ref{subsec:models}\footnote{Code and hyperparameters used are shared in \url{https://anonymous.4open.science/r/vicarious/}}.

\subsection{Experiment 1: Inter-Group Disagreement} 
\label{sec:exp1}

\textbf{RQ1 (Inter-group disagreement)}: \textit{Do LLM personas replicate patterns of disagreement observed between human demographic groups when judging hate speech?}

\paragraph{Setup.}  
For each demographic pair $(A, B)$, we evaluate inter-group disagreement on the subset of comments labelled by both groups~(see Appendix Table~\ref{tab:data_rq1_overlap_comments}). Humans provide majority-vote labels per group, while LLM personas generate a single label per comment. Only overlapping comment sets are used to compute agreement (see Section~\ref{subsubsec:disagreement_metrics} for metric details).

\begin{table}[h]
    \centering
    \small
    \resizebox{\linewidth}{!}{%
    \begin{tabular}{l r rr rr rr}
    \toprule
    & & \multicolumn{2}{c}{\texttt{Llama~3.1}} & \multicolumn{2}{c}{\texttt{Nemo}} & \multicolumn{2}{c}{\texttt{Qwen~3}} \\
    \cmidrule(lr){3-4} \cmidrule(lr){5-6} \cmidrule(lr){7-8}
    \textbf{Axis}
    & Human
    & Persona & Vanilla
    & Persona & Vanilla
    & Persona & Vanilla \\
    \midrule
    \textit{Gender}    & 0.649 & 0.258 & 0.496 & \textbf{0.678} & 0.532 & 0.135 & -0.112 \\
    \textit{Race}      & 0.760 & 0.512 & 0.532 & \textbf{0.707} & 0.648 & 0.262 & -0.231 \\
    \textit{Religion}  & 0.645 & 0.000 & 0.573 & \textbf{0.673} & 0.642 & 0.301 & -0.199 \\
    \textit{Ideology}  & 0.812 & 0.209 & 0.704 & 0.328 & \textbf{0.705} & 0.395 & -0.121 \\
    \textit{Sexuality} & 0.863 & 0.550 & 0.728 & \textbf{0.884} & 0.651 & 0.292 & -0.193 \\
    \bottomrule
    \end{tabular}%
    }
    \caption{\textbf{RQ1}: Cross-group Cohen's $\kappa$ for human annotators and LLM personas across demographic axes. Inter-annotator agreement is computed between paired demographic groups for each axis (Men vs.\ Women, White vs.\ Black, Christian vs.\ Atheist, Liberal vs.\ Conservative, and Straight vs.\ Gay). For humans, each comment is assigned the majority-vote label per group; for LLMs, the Persona columns report agreement between identity-conditioned predictions per group, while the Vanilla columns report agreement of a single non-persona baseline prediction.}
    \label{tab:exp1_results}
\end{table}

\paragraph{Results.}
Table~\ref{tab:exp1_results} reports cross-group $\kappa$ for humans and LLM personas across five demographic axes, computed on overlapping comment sets. Human annotators show consistently high agreement (0.645–0.863), indicating substantial consensus across demographic pairs. 

LLM performance varies across models and axes. \texttt{Nemo} approximates human agreement on Gender, Race, Religion, and Sexuality, but diverges on Ideology (0.328 vs.\ 0.812). \texttt{Llama~3.1} shows low agreement across all axes, including zero agreement on Religion, while \texttt{Qwen~3} remains low across the board (0.135–0.395).   

The non-persona baseline (Vanilla) shows that adding demographic personas can actually hurt alignment with human judgments. For \texttt{Llama~3.1}, the vanilla setting already aligns reasonably well with humans (0.496–0.728), but persona prompting sharply reduces agreement, dropping to zero on Religion. \texttt{Nemo} behaves differently: personas generally match or slightly improve on the baseline, except for Ideology, where agreement falls from 0.705 to 0.328 when split into Liberal vs. Conservative. In contrast, \texttt{Qwen~3} starts off misaligned in its non-persona version (negative agreement), but persona prompting partially corrects this, bringing scores into a positive range (0.135–0.395), though still below the other models.

These results suggest that some LLMs, particularly \texttt{Nemo}, can reproduce human cross-group patterns on most axes, though certain dimensions, such as ideology, remain challenging. 

To assess sensitivity to subset size, we evaluated the two smallest axes at half their original size (Ideology: $n=48$, Sexuality: $n=51$) over 1{,}000 iterations and compared the resulting Cohen’s $\kappa$ to the full-sample estimates. For \textit{Ideology} ($n=96$), agreement remains effectively unchanged: \texttt{Llama~3.1} (0.209 $\rightarrow$ 0.211), \texttt{Nemo} (0.328 $\rightarrow$ 0.324), and \texttt{Qwen~3} (0.395 $\rightarrow$ 0.394). A similar pattern holds for \textit{Sexuality} ($n=102$): \texttt{Llama~3.1} (0.550 $\rightarrow$ 0.549), \texttt{Nemo} (0.884 $\rightarrow$ 0.881), and \texttt{Qwen~3} (0.292 $\rightarrow$ 0.295). Across both axes, half-size estimates closely match the full-sample $\kappa$ values, with overlapping confidence intervals in all cases. These results indicate that, although scores for smaller subsets such as \textit{Ideology} and \textit{Sexuality} should be interpreted with caution, the relative ordering of the models remains stable.

Having established patterns of inter-group agreement, we next examine whether LLM personas also capture finer-grained effects, such as increased sensitivity to comments targeting their own identity.

\begin{table*}[h]
    \centering
    \small
    \resizebox{\linewidth}{!}{%
    \begin{tabular}{ll | rrr | rrr rrr rrr}
        \toprule
        \multirow{2}{*}{\textbf{Axis}} & \multirow{2}{*}{\textbf{Persona}}
        & \multicolumn{3}{c}{\textbf{Human}}
        & \multicolumn{3}{c}{\texttt{Llama~3.1}}
        & \multicolumn{3}{c}{\texttt{Nemo}} 
        & \multicolumn{3}{c}{\texttt{Qwen~3}} \\
        \cmidrule(lr){3-5} \cmidrule(lr){6-8} 
        \cmidrule(lr){9-11} \cmidrule(lr){12-14}
         &   & $\kappa^{\text{targ}}$ & $\kappa^{\text{non-targ}}$ 
        & $\Delta_{\text{IG}}$
        & $\kappa^{\text{targ}}$ & $\kappa^{\text{non-targ}}$ 
        & $\Delta_{\text{IG}}$
        & $\kappa^{\text{targ}}$ & $\kappa^{\text{non-targ}}$ 
        & $\Delta_{\text{IG}}$
        & $\kappa^{\text{targ}}$ & $\kappa^{\text{non-targ}}$ 
        & $\Delta_{\text{IG}}$ \\
        \midrule
        \multirow{2}{*}{\textit{Gender}}
        & Men   & 0.389 & 0.537 & -0.148\,* &  0.126 &  0.187 & -0.062\,* & -0.032 & -0.024 & -0.008\,* &  0.005 &  0.019 & -0.014\,\phantom{*} \\
        & Women & 0.370 & 0.526 & -0.156\,* &  0.032 &  0.031 & +0.001\,* & -0.038 & -0.019 & -0.019\,* &  0.005 & -0.009 & +0.014\,* \\
        \midrule
        \multirow{2}{*}{\textit{Race}}
        & White & 0.307 & 0.511 & -0.205\,* &  0.251 &  0.165 & +0.086\,\phantom{*} & -0.030 & -0.044 & +0.013\,\phantom{*} & -0.042 &  0.054 & -0.096\,* \\
        & Black & 0.669 & 0.473 & +0.195\,* &  0.272 &  0.192 & +0.080\,* & -0.009 & -0.030 & +0.021\,* & -0.083 & -0.037 & -0.046\,* \\
        \midrule
        \multirow{2}{*}{\textit{Religion}}
        & Christian & 0.262 & 0.515 & -0.253\,\phantom{*} & -0.006 &  0.004 & -0.011\,* & -0.025 & -0.038 & +0.013\,* &  0.011 &  0.009 & +0.003\,\phantom{*} \\
        & Atheist   & 0.130 & 0.512 & -0.383\,\phantom{*} & -0.050 &  0.089 & -0.139\,\phantom{*} & -0.097 & -0.044 & -0.053\,\phantom{*} & -0.009 &  0.009 & -0.018\,\phantom{*} \\
        \midrule
        \multirow{2}{*}{\textit{Sexuality}}
        & Straight & -0.200 & 0.533 & -0.733\,\phantom{*} &  0.333 &  0.423 & -0.089\,\phantom{*} &  0.000 & -0.022 & +0.022\,\phantom{*} & -0.059 &  0.023 & -0.082\,\phantom{*} \\
        & Gay      &  0.685 & 0.478 & +0.207\,\phantom{*} &  0.531 &  0.202 & +0.329\,\phantom{*} & -0.099 & -0.034 & -0.065\,* &  0.010 & -0.028 & +0.038\,* \\
        \bottomrule
    \end{tabular}
    }
    \caption{\textbf{RQ2}: In-group sensitivity across demographic groups and model families. $\kappa^{\text{targ}}$ and $\kappa^{\text{non-targ}}$ report Cohen's $\kappa$ for Targeted (comments directed at group $A$) and Non-targeted (comments not directed at $A$) conditions. For humans, $\kappa$ reflects agreement between group $A$'s majority vote and the remaining annotators ($\neg A$). For LLM personas, $\kappa$ reflects agreement between the persona's predictions and the remaining annotators ($\neg A$). $\Delta_{\text{IG}} = \kappa^{\text{targ}} - \kappa^{\text{non-targ}}$ indicates the change in agreement: positive values mean higher alignment when the group's own identity is targeted. * denotes $p < 0.05$ after Benjamini--Hochberg correction.}
    \label{tab:rq2}
\end{table*}

\subsection{Experiment 2: In-Group Sensitivity} 
\label{sec:exp2}

\textbf{RQ2 (In-group sensitivity)}: \textit{Do annotators judge hate speech differently when their own identity group is targeted, and do LLM personas exhibit similar patterns relative to human annotations?}

\paragraph{Setup.}
For each group $A$, we evaluate in-group sensitivity on two subsets of comments: Targeted (comments explicitly targeting group $A$) and Non-targeted (all other comments; see Appendix Table~\ref{tab:data_rq2_targeted_comments}). Agreement is assessed using the in-group sensitivity metric $\Delta_{\text{IG}}$ defined in Section~\ref{subsec:in-group-sensitivity}.

\paragraph{Results.}
Table~\ref{tab:rq2} reports in-group sensitivity ($\Delta_{\text{IG}}$) for human annotators and LLM personas across Gender, Race, Religion, and Sexuality.\footnote{No comments explicitly target the Ideology axis, so it is excluded from this analysis.} 

Human annotators show systematic changes in agreement with other annotators depending on whether comments target their own identity. For example, Black ($\Delta_{\text{IG}} = +0.195$) and Gay ($+0.207$) annotators show increased agreement with the out-group on comments targeting their own identity, while White ($-0.205$), Men ($-0.148$), and Women ($-0.156$) show decreased agreement on such comments.

LLM personas exhibit mixed patterns. \texttt{Llama~3.1} shows positive boosts for Black ($+0.080$) and Gay ($+0.329$) personas, consistent in direction with the corresponding human groups. For most other groups, however, Llama's $\Delta_{\text{IG}}$ values are near zero or negative, indicating limited sensitivity to whether comments target the persona's identity. The Atheist persona shows a negative boost ($-0.139$) that diverges from the human pattern. \texttt{Nemo} and \texttt{Qwen~3} show mostly near-zero or negative shifts across axes, with some values moving in the opposite direction to human trends (e.g., Race personas for Qwen~3: White $-0.096$, Black $-0.046$).  

Overall, only \texttt{Llama~3.1} partially reproduces human in-group sensitivity for certain groups. No model consistently captures the directional pattern of human agreement changes across demographic axes.  

These results set the stage for the next analysis, which examines whether personas can adopt an out-group perspective and predict how a different demographic would judge the same content.

\subsection{Experiment 3: Out-Group Perception} 
\label{sec:exp3}

\textbf{RQ3 (Out-group perception)}: \textit{Do LLM personas change their hate speech judgements under vicarious prompting (i.e., when evaluating from another group’s perspective), and how do these predictions align with human consensus?}

\paragraph{Setup.}
For each demographic pair $(A, B)$, persona $A$ evaluates comments under two conditions: self-alignment, rating from its own perspective, and vicarious, predicting how group $B$ would label the same comment. For example, on the Gender axis, a prompt may take the form: \textit{``You are a man. Would women consider the following comment to be hate speech?''} (see~\ref{subsec:prompting} for exact prompts). Statistical analyses and metric definitions are described in Section~\ref{subsec:metrics}.

\begin{table*}[t]
    \centering
    \resizebox{\linewidth}{!}{%
    \begin{tabular}{ll
    ccc ccc
    ccc ccc
    ccc ccc}
    \toprule
    
    & &
    \multicolumn{6}{c}{\texttt{Llama~3.1}} &
    \multicolumn{6}{c}{\texttt{Nemo}} &
    \multicolumn{6}{c}{\texttt{Qwen~3}} \\
    
    \cmidrule(lr){3-8} \cmidrule(lr){9-14} \cmidrule(lr){15-20}
    
    & &
    \multicolumn{3}{c}{Out-group} &
    \multicolumn{3}{c}{Global} &
    \multicolumn{3}{c}{Out-group} &
    \multicolumn{3}{c}{Global} &
    \multicolumn{3}{c}{Out-group} &
    \multicolumn{3}{c}{Global} \\
    
    \cmidrule(lr){3-5} \cmidrule(lr){6-8}
    \cmidrule(lr){9-11} \cmidrule(lr){12-14}
    \cmidrule(lr){15-17} \cmidrule(lr){18-20}
    
    \textbf{Axis} &
    \textbf{Observer $\rightarrow$ OG}
    
    & $\kappa^{\text{self}}$ & $\kappa^{\text{vic}}$ & $\Delta$
    & $\kappa^{\text{self}}$ & $\kappa^{\text{vic}}$ & $\Delta$
    
    & $\kappa^{\text{self}}$ & $\kappa^{\text{vic}}$ & $\Delta$
    & $\kappa^{\text{self}}$ & $\kappa_{\text{vic}}$ & $\Delta$
    
    & $\kappa^{\text{self}}$ & $\kappa^{\text{vic}}$ & $\Delta$
    & $\kappa^{\text{self}}$ & $\kappa^{\text{vic}}$ & $\Delta$ \\
    \midrule
    
    \multirow{2}{*}{\textit{Gender}}
    & Men $\rightarrow$ Women
    & 0.173 & 0.441 & +0.268* & 0.173 & 0.372 & +0.199*
    & -0.033 & -0.032 & +0.001* & -0.028 & -0.024 & +0.004*
    & 0.041 & 0.015 & -0.026 & 0.015 & 0.010 & -0.005* \\
    
    & Women $\rightarrow$ Men
    & 0.033 & 0.425 & +0.392* & 0.035 & 0.391 & +0.356*
    & -0.025 & -0.062 & -0.037* & -0.028 & -0.040 & -0.012*
    & 0.006 & 0.052 & +0.046* & -0.006 & 0.047 & +0.053* \\
    
    \midrule
    
    \multirow{2}{*}{\textit{Race}}
    & White $\rightarrow$ Black
    & 0.168 & 0.466 & +0.298* & 0.145 & 0.430 & +0.285*
    & -0.100 & -0.085 & +0.015* & -0.044 & -0.044 & 0.000*
    & 0.089 & -0.012 & -0.101\phantom{*} & 0.035 & -0.023 & -0.058\phantom{*} \\
    
    & Black $\rightarrow$ White
    & 0.218 & 0.473 & +0.255* & 0.221 & 0.424 & +0.203\phantom{*}
    & -0.016 & -0.020 & -0.004* & -0.023 & -0.032 & -0.009*
    & -0.051 & 0.008 & +0.059\phantom{*} & -0.035 & 0.007 & +0.042\phantom{*} \\
    
    \midrule
    
    \multirow{2}{*}{\textit{Religion}}
    & Chr. $\rightarrow$ Atheist
    & 0.004 & 0.387 & +0.384\phantom{*} & 0.003 & 0.342 & +0.339*
    & -0.020 & -0.042 & -0.021* & -0.040 & -0.038 & +0.002*
    & 0.044 & 0.031 & -0.012\phantom{*} & 0.000 & 0.024 & +0.024\phantom{*} \\
    
    & Atheist $\rightarrow$ Chr.
    & 0.081 & 0.399 & +0.318* & 0.084 & 0.312 & +0.228*
    & -0.068 & -0.083 & -0.014* & -0.039 & -0.058 & -0.019*
    & -0.005 & 0.015 & +0.021\phantom{*} & 0.008 & 0.003 & -0.005\phantom{*} \\
    
    \midrule
    
    \multirow{2}{*}{\textit{Ideology}}
    & Liberal $\rightarrow$ Con.
    & 0.502 & 0.509 & +0.007\phantom{*} & 0.377 & 0.375 & -0.002*
    & -0.085 & -0.123 & -0.038\phantom{*} & -0.016 & -0.040 & -0.024*
    & 0.018 & -0.069 & -0.087\phantom{*} & -0.017 & 0.004 & +0.021\phantom{*} \\
    
    & Con. $\rightarrow$ Liberal
    & 0.169 & 0.462 & +0.293* & 0.148 & 0.397 & +0.249*
    & -0.075 & -0.103 & -0.028* & -0.025 & -0.028 & -0.003*
    & 0.081 & -0.033 & -0.113* & 0.052 & -0.001 & -0.053\phantom{*} \\
    
    \midrule
    
    \multirow{2}{*}{\textit{Sexuality}}
    & Straight $\rightarrow$ Gay
    & 0.492 & 0.462 & -0.030\phantom{*} & 0.370 & 0.376 & +0.006*
    & -0.019 & -0.060 & -0.042\phantom{*} & -0.032 & -0.035 & -0.003*
    & 0.042 & 0.149 & +0.107\phantom{*} & 0.022 & 0.012 & -0.010\phantom{*} \\
    
    & Gay $\rightarrow$ Straight
    & 0.344 & 0.434 & +0.091\phantom{*} & 0.293 & 0.379 & +0.086\phantom{*}
    & -0.042 & -0.045 & -0.002* & -0.026 & -0.024 & +0.002*
    & -0.010 & -0.013 & -0.003\phantom{*} & -0.003 & 0.002 & +0.005\phantom{*} \\
    
    \bottomrule
    \end{tabular}}
    \caption{\textbf{RQ3}: Out-group perception across demographic pairs and model families. For each observer $\rightarrow$ out-group direction, predictions are evaluated against two reference populations: \textit{Out-group} (OG) compares against the majority vote of the target demographic group, while \textit{Global} compares against the majority vote of the entire dataset. $\kappa^{self}$ and $\kappa^{vic}$ denote Cohen's $\kappa$ for self-alignment and vicarious predictions, respectively. $\Delta = \kappa^{vic} - \kappa^{self}$ captures the vicarious gap: positive values indicate that adopting the out-group's perspective moves predictions closer to that group's actual annotations, while negative values indicate the opposite. * denotes $p < 0.05$ after Benjamini--Hochberg correction.}
    \label{tab:rq3}
\end{table*}

\paragraph{Results.}
Table~\ref{tab:rq3} reports the vicarious gap $\Delta$ across demographic pairs and model families, evaluated against the out-group majority vote. Positive $\Delta$ indicates that vicarious prompting moves predictions closer to the out-group's annotations than the persona's own ratings.

\texttt{Llama~3.1} shows consistent positive $\Delta$ across most axes, with the largest effects for Gender (Women $\rightarrow$ Men: $+0.392$; Men $\rightarrow$ Women: $+0.268$), Race (White $\rightarrow$ Black: $+0.298$; Black $\rightarrow$ White: $+0.255$), and Religion (Atheist $\rightarrow$ Christian: $+0.318$). In contrast, \texttt{Nemo} shows mostly negative or near-zero $\Delta$, and \texttt{Qwen~3} has low agreement overall.

When comparing predictions against the global majority vote (Global columns), \texttt{Llama~3.1} again shows mostly positive gaps, indicating that adopting an out-group perspective not only improves alignment with specific groups but also increases agreement with the overall human consensus. The largest global shifts occur for Women $\rightarrow$ Men ($+0.356$) and White $\rightarrow$ Black ($+0.285$), consistent with the out-group analysis. \texttt{Nemo} and \texttt{Qwen~3} remain inconsistent, with most gaps non-significant, confirming that perspective-taking is model-dependent. 

Overall, \texttt{Llama~3.1} is the only model that reliably benefits from vicarious prompting, improving alignment with both target groups and the global majority.

Having evaluated each dimension independently, we next compare the best-performing configuration from each experiment on a shared subset of comments in order to assess how these approaches differ when evaluated under a common setting.

\subsection{Summary}
We directly compared the three best settings across identical subsets of comments and demographic groups, alongside the best non-persona baseline, to evaluate whether identity conditioning improves over plain prompting. Specifically, we restricted the evaluation to the data subset and comparison settings used in RQ1. The selected configurations were:

\begin{enumerate}
    \item \textbf{RQ1 \texttt{Nemo} with \textit{Self-Alignment} prompt:} showed the highest human-level agreement within each demographic.
    \item \textbf{RQ2 \texttt{Llama~3.1} with \textit{Self-Alignment} prompt:} captured in-group sensitivity to comments targeting the persona's own identity.
    \item \textbf{RQ3 \texttt{Llama~3.1} with \textit{Vicarious} prompt:} accurately predicted out-group perceptions, effectively performing perspective-taking.
    \item \textbf{\texttt{Nemo} with \textit{non-persona} prompt:} a plain prompt with no identity conditioning, serving as a reference for label quality without demographic framing. We use \texttt{Nemo} non-persona as the baseline model in this summary, since it yielded the highest cross-group agreement without persona prompting in Section~\ref{sec:exp1}.
\end{enumerate}

\begin{table}[h]
    \centering
    \small
    \resizebox{\linewidth}{!}{%
    \begin{tabular}{lccccc}
    \toprule
    \textbf{Dem.} & \textbf{Human $\kappa$} & \textbf{Nemo Vanilla $\kappa$} & \textbf{Nemo $\kappa_{\text{self}}$} & \textbf{Llama $\kappa_{\text{self}}$} & \textbf{Llama $\kappa_{\text{vic}}$} \\
    \midrule
    Gender     & 0.649 & 0.532 & 0.678 & 0.258 & \textbf{0.861} \\
    Race       & 0.760 & 0.648 & 0.707 & 0.512 & \textbf{0.809} \\
    Religion   & 0.645 & 0.642 & 0.673 & 0.000 & \textbf{0.732} \\
    Ideology   & 0.812 & 0.705 & 0.328 & 0.209 & \textbf{0.710} \\
    Sexuality  & 0.863 & 0.651 & \textbf{0.884} & 0.550 & 0.803 \\
    \bottomrule
    \end{tabular}
    }
    \caption{Cross-group Cohen's $\kappa$ for human annotators and LLM configurations across demographic axes. For humans, each comment is assigned the majority-vote label per group; for individual LLM personas, $\kappa$ is computed between the two demographic groups (e.g., Men vs.\ Women). Bold indicates the highest LLM-persona $\kappa$ per demographic axis.}
    \label{tab:summary}
\end{table}

Across demographic axes, \texttt{Llama~3.1} with vicarious prompting achieves the highest cross-group agreement in four of the five categories (gender, race, religion, and ideology), and \texttt{Nemo} with persona-self prompting, obtains the highest agreement for sexuality. In contrast, \texttt{Llama~3.1} self-alignment shows substantially lower agreement across all axes, including zero agreement for religion. The non-persona baseline (Vanilla) provides a useful reference: it reaches moderate agreement with the human majority (0.532--0.705). \texttt{Nemo} with self-alignment improves over this baseline on all axes except ideology, suggesting that its persona prompting adds meaningful demographic signal. In contrast, \texttt{Llama~3.1} self-alignment underperforms the non-persona prompting on most axes, indicating that its personas introduce divergence that does not reflect human patterns. \texttt{Llama~3.1} with vicarious prompting is the only setup that consistently exceeds both the vanilla setting and human cross-group $\kappa$ on four of five axes, highlighting perspective-taking as the most effective strategy for demographically-aware annotation. However, exceeding human $\kappa$ does not necessarily indicate better performance: it may reflect a flattening of genuine inter-group differences rather than improved consistency. By this criterion, \texttt{Nemo} with self-alignment, whose $\kappa$ values track human cross-group agreement most closely across all five axes, may represent the most calibrated setting for tasks where preserving demographic signal is desirable.

\section{Discussion}
Our results across RQ1–RQ3 show that persona-conditioned judgement patterns are highly model-dependent and do not reliably emerge from minimal identity prompts alone. This finding is consistent with \citet{baumann2025largelanguagemodelhacking}, who report that model selection is the primary driver of variability in LLM annotation outputs, and with \citet{hu2025simbenchbenchmarkingabilitylarge}, who demonstrate that even state-of-the-art models show limited fidelity when simulating human behaviour across demographic groups. Together, these results suggest that the apparent social reasoning capacity of persona-conditioned LLMs primarily reflects model-specific training distributions and alignment regimes rather than a general ability to adopt demographic perspectives.

Despite the limitations of self-alignment prompting, vicarious prompting with \texttt{Llama~3.1} produces cross-group agreement that matches human cross-group $\kappa$ on four out of five demographic axes (Table~\ref{tab:summary}). This is a notable result: rather than asking a model to act as a demographic group, asking it to predict how that group would respond yields substantially higher cross-group agreement. One possible explanation, which we leave for future empirical investigation, is related to the False Consensus Effect in LLMs described by \citet{choi-etal-2025-people}: self-alignment prompts may lead models to project their default judgements onto the persona, whereas vicarious prompts explicitly require perspective-taking, which may partially counteract this bias. It should be noted, however, that $\kappa$ values exceeding human cross-group agreement do not straightforwardly indicate superior performance, as they may reflect a suppression of genuine demographic variation rather than improved consistency.

These findings have direct implications for the use of LLM personas in hate speech annotation. On the one hand, the strong cross-group agreement observed for \texttt{Llama~3.1} with vicarious prompting suggests that this strategy may help approximate patterns of cross-group consistency in situations where collecting annotations from specific demographic groups is costly or infeasible. On the other hand, as \citet{moscato-etal-2025-personalization} caution, moderation systems that uncritically rely on simulated group preferences risk misrepresenting minority perspectives and violating legal hate speech boundaries. This risk is further compounded by \citet{sap-etal-2022-annotators}, who show that annotation biases tied to race and ideology are systematic and directional, meaning that errors introduced by imperfect persona simulation are unlikely to cancel out randomly.

In practical terms, Table~\ref{tab:summary} shows that \texttt{Llama~3.1} vicarious prompting achieves $\kappa^{\text{vic}}$ values that match or exceed human cross-group $\kappa$ for Gender (0.861 vs. 0.649), Race (0.809 vs. 0.760), Religion (0.732 vs. 0.645), and approach it for Ideology (0.710 vs. 0.812), and Sexuality (0.803 vs. 0.863). Nemo's self-alignment prompts also match on Gender, Race, Religion and Sexuality, while Llama's self-alignment generally underperforms except on Race and Sexuality, where it shows only a moderate agreement. Taken together, these results suggest that the optimal configuration depends on the annotation objective: vicarious prompting maximises cross-group consistency, while \texttt{Nemo} self-alignment may be preferable when closer calibration to human-level demographic signals is desired. 

We therefore echo the recommendation of \citet{baumann2025largelanguagemodelhacking}: LLM-based annotations should be treated as instruments requiring validation rather than convenient black-box substitutes for human judgement. Participatory evaluation with members of the relevant communities remains essential before deployment, particularly when annotations inform sensitive moderation or research tasks.

\section{Conclusion}
\label{sec:conclusion}
This study evaluated whether zero-shot persona-conditioned LLMs reproduce structural properties of human judgement in hate speech perception. Across three model families and five demographic axes, model behaviour varies by model and evaluation dimension. On inter-group disagreement (RQ1), \texttt{Nemo} best approximates human cross-group agreement, while \texttt{Llama~3.1} and \texttt{Qwen~3} show lower inter-persona agreement. On in-group sensitivity (RQ2), only \texttt{Llama~3.1} partially replicates the human pattern for Black and Gay personas. On out-group perception (RQ3), only \texttt{Llama~3.1} with vicarious prompting produces judgements closer to the out-group's labels.

Taken together, these findings show that, under minimal zero-shot persona prompting, LLMs do not consistently reproduce the social structure of human hate speech perception. Some models capture specific dimensions of human judgement, but no configuration generalises across all three. Among the settings tested, vicarious prompting with \texttt{Llama~3.1} most consistently approximates human cross-group disagreement patterns, suggesting that perspective-taking prompts may provide a more reliable basis for automated annotation. However, model alignment with human judgement remains strongly model-dependent, indicating that persona-based annotation pipelines should be empirically validated rather than assumed to faithfully represent demographic perspectives

These results reflect zero-shot persona conditioning on an English dataset and may improve with stronger prompts, calibration, or fine-tuning. Future work should explore whether richer persona specifications, chain-of-thought prompting, or model variants affect these findings. It should also test whether they generalise to non-English datasets. Finally, researchers could investigate whether a jury of vicarious LLM personas can further improve cross-group consistency.

\section*{Limitations}
This study has several limitations. First, the analysis focuses exclusively on zero-shot prompting in order to capture models' baseline behaviour. While alternative strategies such as few-shot prompting or instruction refinement may improve alignment with human judgements, evaluating such interventions falls outside the scope of this work and represents an important direction for future research. Second, the study relies on the MHS dataset, whose annotations were produced primarily by US-based annotators. Consequently, the observed disagreement patterns reflect the cultural context of this population and may not generalise fully to other sociocultural settings. Future work should examine datasets with broader geographic and cultural coverage. Third, our analysis simplifies demographic comparisons into binary group pairs to maximise statistical reliability, which necessarily reduces the intersectional complexity of real-world identities. Finally, the models evaluated represent only three open LLM families, and the behaviours identified may differ in other architectures or more heavily aligned systems.

\section*{Ethical Considerations}
Persona-conditioned LLMs show promise for exploring subjective perspectives in hate speech annotation, but their outputs are highly model-dependent and do not consistently reflect the full diversity of human judgements. While some models capture inter-group disagreement or perspective-taking in specific contexts, others diverge. Using LLM personas as proxies for human annotators carries the risk of misrepresentation or overgeneralisation if applied without validation. We therefore recommend treating these outputs as complementary to, rather than replacements for, human judgement, and stress careful empirical evaluation before deploying them in moderation or social science contexts.

\section*{Acknowledgments}
The authors thank the funding from the Horizon Europe research and innovation programme under the Marie Skłodowska-Curie Grant Agreement No. 101073351.  The authors also thank the financial support supplied by the grant PID2022-137061OB-C21 funded by MI-CIU/AEI/10.13039/501100011033 and by “ERDF/EU”. The authors also thank the funding supplied by the Consellería de Cultura, Educación, Formación Profesional e Universidades (accreditations ED431G 2023/01 and ED431C 2025/49) and the European Regional Development Fund, which acknowledges the CITIC, as a center accredited for excellence within the Galician University System and a member of the CIGUS Network, receives subsidies from the Department of Education, Science, Universities, and Vocational Training of the Xunta de Galicia. Additionally, it is co-financed by the EU through the FEDER Galicia 2021-27 operational program (Ref. ED431G 2023/01).

\bibliography{custom}

\appendix
\section{Appendix}
\label{sec:appendix}

\subsection{Data}

\begin{table}[h]
    \centering
    \small
    \resizebox{\linewidth}{!}{
    \begin{tabular}{l l r r}
    \toprule
    \textbf{Dem. Axis} & \textbf{Group} & \textbf{Annot.} & \textbf{Comments} \\
    \midrule
    
    \multirow{4}{*}{\textit{Gender}}
    & Women & 70,699 & 26,108 \\
    & Men & 53,279 & 22,786 \\
    & Non-binary & 932 & 670 \\
    & Prefer not to say & 456 & 358 \\
    
    \midrule
    \multirow{7}{*}{\textit{Race}}
    & White & 92,992 & 28,407 \\
    & Black & 12,208 & 7,376 \\
    & Asian & 8,834 & 5,443 \\
    & Latinx & 8,094 & 5,046 \\
    & Native American & 1,623 & 1,138 \\
    & Middle Eastern & 589 & 437 \\
    & Pacific Islander & 268 & 229 \\
    
    \midrule
    \multirow{8}{*}{\textit{Religion}}
    & Christian & 53,655 & 22,702 \\
    & Nothing & 33,322 & 16,633 \\
    & Atheist & 25,330 & 13,429 \\
    & Buddhist & 1,872 & 1,270 \\
    & Jewish & 1,774 & 1,191 \\
    & Mormon & 892 & 659 \\
    & Muslim & 848 & 617 \\
    & Hindu & 550 & 426 \\

    \midrule
    \multirow{6}{*}{\textit{Ideology}}
    & Liberal & 31,237 & 15,962 \\
    & Neutral & 21,491 & 11,907 \\
    & Slightly Liberal & 19,821 & 11,112 \\
    & Conservative & 14,340 & 8,524 \\
    & Slightly Conservative & 13,977 & 8,490 \\
    & Extremely Conservative & 4,162 & 2,694 \\

    \midrule
    \multirow{4}{*}{\textit{Sexuality}}
    & Straight & 106,494 & 29,194 \\
    & Bisexual & 11,400 & 6,868 \\
    & Gay & 4,946 & 3,177 \\
    & Other & 1,745 & 1,195 \\
    
    \bottomrule
    \end{tabular}
    }
    \caption{Demographic distribution of the \textit{Measuring Hate Speech} dataset. ``Annot.'' refers to the total number of labels provided by annotators of that group. ``Comments'' refers to the number of distinct comments rated by at least one member of that group.}
    \label{tab:full_demographics}
\end{table}

\paragraph{RQ1 Data Subset.}
Table~\ref{tab:data_rq1_overlap_comments} reports the number of comments for each demographic axis. These counts correspond to the subset of comments labelled by both groups with at least two annotators per group. Because not all groups rated the same comments, this intersection defines the comment sets on which Cohen's $\kappa$ is computed, ensuring a reliable measure of cross-group agreement.

\begin{table}[h]
    \centering
    \small
    \resizebox{\linewidth}{!}{
    \begin{tabular}{l l l r}
    \toprule
    \textbf{Dem. Axis} & \textbf{Group A} & \textbf{Group B} & \textbf{\#} \\
    \midrule
    Gender       & Women & Men          & 1,697 \\
    Race         & White & Black        & 213 \\
    Religion     & Christian & Atheist  & 304 \\
    Ideology     & Liberal & Conservative & 96 \\
    Sexuality    & Straight & Gay       & 102 \\
    \bottomrule
    \end{tabular}
    }
    \caption{Number of overlapping comments per demographic axis. These correspond to the comments with at least two annotations from each group.}
    \label{tab:data_rq1_overlap_comments}
\end{table}

\paragraph{RQ2 Data Subset.}
Table~\ref{tab:data_rq2_targeted_comments} reports the number of comments targeting each demographic group that were evaluated by annotators from that group (\textit{Targeted}), along with the number of comments evaluated by the same group where that group is not the target (\textit{Non-Targeted}). The Non-Targeted set, therefore, includes all other comments annotated by the group in which the group itself is not targeted, including comments that target other groups or that contain no explicit target. For example, there are 2,970 comments targeting women annotated by women, and 13,610 comments annotated by women where women are not the target

\begin{table}[h]
    \centering
    \small
    \resizebox{\linewidth}{!}{
    \begin{tabular}{l l r r}
    \toprule
    \textbf{Dem. Axis} & \textbf{Group} & \textbf{\# Targeted} & \textbf{\# Non-Targeted} \\
    \midrule
    \multirow{2}{*}{\textit{Gender}}
    & Men     & 4,610  & 11,680 \\
    & Women   & 2,970  & 13,610 \\
    \midrule
    \multirow{2}{*}{\textit{Race}}
    & White   & 287    & 5,539  \\
    & Black   & 610    & 5,178  \\
    \midrule
    \multirow{2}{*}{\textit{Religion}}  
    & Christian & 318  & 7,650  \\
    & Atheist   & 38   & 8,017  \\
    \midrule
    \multirow{2}{*}{\textit{Sexuality}} 
    & Straight  & 16   & 2,666  \\
    & Gay       & 216  & 2,423  \\
    \bottomrule
    \end{tabular}
    }
    \caption{Number of comments targeting each group (Targeted) and comments not targeting that group (Non-Targeted).}
    \label{tab:data_rq2_targeted_comments}
\end{table}

\paragraph{RQ3 Data Subset.}
Table~\ref{tab:data_rq3} summarises the subset sizes used to compute vicarious and self predictions. For each demographic pair and observer–out-group combination, it reports the number of predictions generated by the LLM under both vicarious and self prompts (\# LLM preds), the number of comments included in the out-group majority vote (\# Out-group), and the total number of comments with at least one human annotation for the demographic (\# Global).

\begin{table}[h]
    \centering
    \small
    \resizebox{\linewidth}{!}{
    \begin{tabular}{ll rrr}
    \toprule
    \textbf{Demographic} & \textbf{Observer $\rightarrow$ Out-group} 
    & \# LLM preds
    & \# Out-group 
    & \# Global \\
    \midrule
    \multirow{2}{*}{\textit{Gender}}    
    & Men $\rightarrow$ Women & 21,846 & 7,414 & 29,339 \\
    & Women $\rightarrow$ Men   & 25,092 & 5,749 & 29,339  \\
    \midrule
    \multirow{2}{*}{\textit{Race}}
    & White $\rightarrow$ Black & 27,791 & 470   & 29,304 \\
    & Black $\rightarrow$ White & 6,874  & 3,173 & 29,304 \\
    \midrule
    \multirow{2}{*}{\textit{Religion}}
    & Christian $\rightarrow$ Atheist & 21,704 & 1,316 & 26,601 \\
    & Atheist $\rightarrow$ Christian & 12,711 & 2,472 & 26,601 \\
    \midrule
    \multirow{2}{*}{\textit{Ideology}}
    & Liberal $\rightarrow$ Conservative & 14,972 & 281 & 26,405 \\
    & Conservative $\rightarrow$ Liberal      & 8,044  & 575 & 26,405 \\
    \midrule
    \multirow{2}{*}{\textit{Sexuality}}
    & Straight $\rightarrow$ Gay & 28,791 & 162  & 29,322 \\
    & Gay $\rightarrow$ Straight & 2,944  & 1,647 & 29,322 \\
    \bottomrule
    \end{tabular}
    }
    \caption{Subset sizes used for $\kappa$ computation in RQ3. For each demographic pair and observer-out-group combination, the table reports: $(1)$ the number of LLM predictions (\# LLM preds) generated with vicarious and self prompts, $(2)$ the number of comments included in the out-group majority vote (``\# Out-group''), and the total number of comments with at least one human annotation for the demographic (``\# Global'').}
    \label{tab:data_rq3}
\end{table}

\subsection{Prompts}
\label{app:prompts}

\subsubsection{Non-Persona}

The model rates a comment in a neutral setting, without adopting any perspective.

\begin{enumerate}
    \item \textbf{Non-Persona:}  
    ``\texttt{Determine whether the following text contains hate speech or not. Respond with either True for hate speech or False for neutral.}''
\end{enumerate}

\subsubsection{Self-Alignment}

The model adopts the identity of demographic group $A$ and rates the comment from that perspective.

\begin{enumerate}
    \item \textbf{Minimal Identity:}  
    ``\texttt{Adopt the identity of [Demographic A]. Answer the questions while staying in strict accordance with the nature of this identity. Determine whether the following text contains hate speech or not based on your experiences and background. Respond with either True for hate speech or False for neutral.}''

    \item \textbf{Contextual Identity:} 
   ``\texttt{You are acting as a person who identifies as [Demographic A]. This identity is a central part of your lived experience, shaping your values, social perspective, and sensitivity to language. When evaluating the text below, draw upon the cultural nuances and historical context associated with being [Demographic A]. Determine if the following text constitutes hate speech from this specific social standpoint. Respond with either True for hate speech or False for neutral.}''
    
    \item \textbf{Biographical Identity:} 
    ``\texttt{You are a [Demographic A]. You have grown up navigating the social dynamics, cultural expectations, and systemic patterns that shape the everyday experiences of people who share your identity. Over the course of your life, you have encountered language used to demean, exclude, or threaten people like you, and you have also witnessed how such language is sometimes dismissed or normalised by those outside your group. Your sensitivity to hate speech has been shaped by these accumulated experiences — what you have heard directed at you or those close to you, what you have seen go unaddressed, and what you know to be harmful even when others do not. With this background in mind, determine whether the following text constitutes hate speech from your perspective. Respond with either True for hate speech or False for neutral.}''
\end{enumerate}

\subsubsection{Vicarious}

The model adopts the identity of demographic group $A$ (Observer) to predict the reaction of group $B$ (Out-group).

\begin{enumerate}
    \item \textbf{Minimal Identity:} 
    ``\texttt{Adopt the identity of [Demographic A]. Your task is not to give your own opinion, but to predict the perception of another person. Based on your understanding of social dynamics, predict whether [Demographic B] would consider the following text to contain hate speech. Respond with either True (if they would see it as hate) or False (if they would not).}''

    \item \textbf{Contextual Identity:}
    ``\texttt{You are acting as [Demographic A]. Your task is to apply your social intelligence and understanding of diverse perspectives to predict how [Demographic B] would feel. Consider the unique sensitivities, lived experiences, and common social reactions of [Demographic B] when they encounter harmful language. Predict whether they would consider the following text to be hate speech. Respond with either True (if they would see it as hate) or False (if they would not).}''
    
    \item \textbf{Biographical Identity:} 
    ``\texttt{You are a [Demographic A]. You have a grounded understanding of your own social position and, through years of observation, conversation, and shared experience, you have developed a genuine awareness of how people from different backgrounds perceive and react to harmful language. You know that what feels neutral or unremarkable to someone in one social position can feel threatening or deeply offensive to someone in another. Your task is not to give your own opinion, but to draw on this social awareness to predict how a [Demographic B] would react to the text below. Consider the specific history of language used to target, demean, or marginalise people with that identity, and the cumulative weight that such language carries for them. Predict whether a [Demographic B] would consider this text to constitute hate speech. Respond with either True (if they would see it as hate) or False (if they would not).}''
\end{enumerate}

\subsection{Robustness and Scaling Analysis}

This section examines whether the findings of Sections~\ref{sec:exp1}-\ref{sec:exp3} hold under variations in model scale and persona prompt richness, using the gender axis as a case study given its largest annotation subset. Moreover, we report the output variance by performing multiple runs.

\subsubsection{Impact of Model Scale}\label{sec:scale}

We analyse scaling effects in \texttt{Llama} by comparing 8B and 70B models on the gender axis, which is the most data-rich and stable demographic setting in our experiments.

For inter-group disagreement (RQ1), human annotators show moderate gender disagreement ($\kappa = 0.649$). The 70B model exceeds this substantially ($\kappa = 0.898$), indicating a collapse of demographic differences, whereas the 8B model underestimates agreement ($\kappa = 0.258$). This suggests a scale-dependent shift from highly divergent to highly convergent persona behaviour.

For  the vicarious prediction (RQ3), at 8B scale, vicarious prompting strongly improves out-group alignment (Men $\rightarrow$ Women: $+0.268$, Women $\rightarrow$ Men: $+0.392$). At 70B scale, effects vanish or become slightly negative (up to $-0.037$), indicating that improved self-alignment reduces the benefit of perspective-taking.

To summarise, scaling improves absolute alignment but reduces sensitivity to persona-based conditioning: smaller models benefit from vicarious prompting, while larger models are largely insensitive to it.

\subsubsection{Sensitivity to Persona Prompting}\label{sec:prompt-sensitivity}

We evaluate three levels of persona prompt richness for \texttt{Llama~3.1~8B}. The \textit{minimal identity} prompt (i.e., the one used across all the experiments) instructs the model to adopt a demographic identity label with no further elaboration. The \textit{contextual identity} prompt frames the identity as a lived experience shaping values and language sensitivity, drawing on cultural and historical context. The \textit{biographical identity} prompt additionally simulates a personal history of exposure to discriminatory language, including experiences of harm being dismissed by out-group members and internalised sensitivity developed over time. Each level adds specificity to the persona without altering the classification task or output format. Exact prompts are reported in Appendix~\ref{app:prompts}.

For RQ1, the \textit{minimal identity} prompt produces the lowest cross-group $\kappa$ ($0.258$), substantially below the human reference ($0.649$). The \textit{contextual identity} prompt narrows this gap considerably ($\kappa = 0.579$, gap $= +0.070$), while the \textit{biographical identity} prompt exceeds the human reference ($\kappa = 0.937$, gap $= -0.288$), indicating that a highly specified persona causes the two demographic personas to converge on nearly identical labels, collapsing the inter-group differentiation that persona conditioning is intended to capture.

For RQ3, the benefit of vicarious prompting decreases steadily with prompt ``richness''. Under the \textit{minimal identity} prompt, vicarious prompting yields large positive gains in both directions (Men$\rightarrow$Women: $\Delta = +0.268$; Women$\rightarrow$Men: $\Delta = +0.392$). Under the \textit{contextual identity} prompt, gains are smaller but remain positive ($\Delta = +0.186$ and $+0.123$). Under the \textit{biographical identity} prompt, vicarious prompting is counterproductive in both directions ($\Delta = -0.191$ and $-0.325$), suggesting that a highly grounded persona resists perspective-taking by over-committing to its own identity framing.

Taken together, these findings mirror the scale results in Section~\ref{sec:scale}: both increasing model size and increasing prompt richness produce personas that agree more internally but deviate further from the human cross-group disagreement structure. The \textit{contextual identity} prompt represents the closest approximation to the human reference on RQ1 ($\kappa = 0.579$) while still benefiting from vicarious prompting on RQ3, suggesting it as the preferable configuration when both inter-group differentiation and perspective-taking are desired. These results also suggest that the relationship between persona specificity and demographic alignment is non-linear: a moderate level of identity grounding improves upon minimal prompts, but excessive biographical detail introduces a rigidity that undermines the model's capacity for social perspective-taking.

\subsubsection{Inference Variance}\label{sec:inference_variance}

Due to computational constraints, we verify empirical stability by running \texttt{Llama~3.1~8B} three times on a subset of gender-axis comments under both self-alignment and vicarious prompting, using a low temperature of $0.2$, and report the mean and standard deviation of Cohen's $\kappa$ across runs. For RQ1, cross-group $\kappa$ remains stable across runs ($\sigma = 0.015$), indicating low sensitivity to stochastic variation in inference. For RQ3, self-alignment and vicarious $\kappa$ values likewise show low variance across runs (men$\rightarrow$women: $\sigma_{\kappa_\text{self}} = 0.021$, $\sigma_{\kappa_\text{vic}} = 0.048$; women$\rightarrow$men: $\sigma_{\kappa_\text{self}} = 0.008$, $\sigma_{\kappa_\text{vic}} = 0.060$), confirming that individual predictions are stable across inference runs.

\subsection{Qualitative Error Analysis}

To complement the quantitative findings, we present a qualitative analysis of illustrative cases drawn from the \textbf{gender} axis, which offers the largest aligned comment set. We identify four patterns for RQ1 and another four for RQ3, examining cases where model behaviour diverges from human annotations.

\subsubsection{RQ1: Inter-Group Disagreement Patterns}

We categorise comments into four scenarios based on the agreement structure between human annotators and the three LLM personas.

\paragraph{Missed Inter-group Disagreement} 
These are comments where men and women \textit{disagreed} in their human annotations, but all three LLM personas produced identical labels for both groups (although can be different among models), failing to capture the genuine inter-group disagreement.

\begin{quote}
\textit{Example:} ``\textbf{She's not a girl she's a fucking whale. Glad to see her get harpooned.}''\\
\small Human annotation: Men = False, Women = True
\end{quote}

\paragraph{True Inter-group Disagreement} 
In very few cases did all three models simultaneously reproduce a human inter-group disagreement. The rarity of this pattern underscores how difficult it is for persona-conditioned LLMs to replicate the social structure of human disagreement under minimal zero-shot prompting.

\begin{quote}
\textit{Example:} ``\textbf{[USER] you're gay, delusional, and retarded for typing that}''\\
\small Humman and LLM annotation: Men = True, Women = False
\end{quote}

\paragraph{Spurious Inter-group Disagreement} 
These comments were judged \textit{identically} by human men and women, yet all three models invented a disagreement that does not exist in the human data. This pattern suggests that identity conditioning can introduce artificial divergence, potentially reflecting stereotypical assumptions about how demographic groups differ rather than grounded perceptual differences.

\begin{quote}
\textit{Example:} ``\textbf{Black and thirty? I'm surprised he wasn't hung by the KKK yet}''\\
\small Human annotation: True $\mid$
Llama: Men = True, Women = False $\mid$
Nemo: Men = False, Women = True $\mid$
Qwen: Men = False, Women = True
\end{quote}

\paragraph{Nemo-Only Success} 
In many cases \texttt{Nemo} successfully differentiated between men and women in a way that matched the human disagreement pattern, while \texttt{Llama~3.1} and \texttt{Qwen~3} both failed to do so. This pattern is consistent with \texttt{Llama}'s partial in-group sensitivity observed in RQ2 and suggests that its training distribution may encode certain gender-related perceptual differences more explicitly.

\begin{quote}
\textit{Example:} ``\textbf{Fuck off cunt. Dont you have some pedophiles to befriend?}''\\
\small Human and Nemo annotation: Men = False, Women = True $\mid$
Llama: True $\mid$ Qwen: False
\end{quote}

\subsubsection{RQ3: Vicarious Perspective-Taking Patterns}

We analyse the men$\rightarrow$women and women$\rightarrow$men vicarious directions separately, categorising comments by the success or failure of the vicarious shift. We used our best vicarious prompting model for this quality analysis (\texttt{Llama~3.1}). We focused on reporting the quality analysis on men$\rightarrow$women (i.e., Men adopted identity, judging as they were women).

\paragraph{Vicarious Correction} 
These are cases where \texttt{Llama~3.1}'s self-alignment prediction was incorrect relative to the women's majority vote, but the vicarious prompt moved the prediction to the correct label. 

\begin{quote}
\textit{Example:} ``\textbf{I will personally rape your mother}''\\
\small Women's human majority = True $\mid$
Llama self (men) = False $\mid$ Llama vicarious (men$\rightarrow$women) = True
\end{quote}

\paragraph{Vicarious Misprediction} 
In these cases \texttt{Llama~3.1}'s self-alignment was \textit{correct}, but the vicarious prompt caused it to shift to the wrong label. This overcorrection pattern is the most prevalent failure mode and suggests that vicarious prompting, while beneficial on average, introduces noise by shifting predictions that did not need to change.

\begin{quote}
\textit{Example:} ``\textbf{Man Chinese people are the worst}''\\
\small Women's human majority = False $\mid$
Llama self (men) = False $\mid$ Llama vicarious (men$\rightarrow$women) = True
\end{quote}

\paragraph{Persistent Misclassification} 
In this scenario, all three models were wrong on self-alignment and none shifted under vicarious prompting. These represent the hardest cases: content where model representations of both the observer and target group are insufficiently calibrated to produce correct labels regardless of prompting strategy.

\begin{quote}
\textit{Example:} ``\textbf{We need another holocaust}''\\
\small Women's human majority = True $\mid$
All models self = False $\mid$ All models vicarious = False
\end{quote}

\paragraph{Llama Vicarious Correction} 
In this last category, \texttt{Llama~3.1}'s vicarious prompt corrected a wrong self-alignment prediction while \texttt{Nemo} and \texttt{Qwen~3} failed to do so even vicariously. This reinforces the finding from RQ3 that perspective-taking effects are strongly model-dependent.

\begin{quote}
\textit{Example:} ``\textbf{Thats nice i hope u end up pregnant bitch and your dad beat his ass now bye.}''\\
\small Women's human majority = True $\mid$
Llama vicarious = True $\mid$ Nemo/Qwen vicarious = False
\end{quote}

\subsection{Generalisation}

While the present study evaluates demographic perspective-taking along binary group axes, a constraint imposed by the annotation density of the MHS dataset, the proposed framework generalises straightforwardly to non-binary settings. For inter-group disagreement (RQ1), pairwise Cohen's $\kappa$ can be extended by computing the average $\kappa$ across all group pairs, yielding a single comparable scalar that scales to any number of groups. For in-group sensitivity (RQ2), the $\neg A$ reference (defined as all annotators outside the focal group) requires no modification, as it naturally accommodates any group partition. For out-group perception (RQ3), practitioners face a design choice: a one-vs-rest formulation, in which group A predicts the reaction of all non-A annotators; an all-pairwise formulation, in which every directed pair (A$→$B, A$→$C, ...) is evaluated separately and gaps are averaged, provides finer-grained insight at greater computational cost. The vicarious prompt template introduced in this work is compatible with both approaches without modification. We therefore encourage practitioners working with richer demographic taxonomies, such as multi-category gender, multi-ethnic race classifications, or intersectional identities, to adopt the pairwise-average or one-vs-rest variants as appropriate to their annotation structure, treating the binary case studied here as a lower bound on the framework's applicability.

\end{document}